# Large Scale Global Optimization by Hybrid Evolutionary Computation

Gutha Jaya Krishna[1, 2] and Vadlamani Ravi[1*]

[1]Center of Excellence in Analytics,
Institute for Development and Research in Banking Technology,
Castle Hills Road #1, Masab Tank, Hyderabad - 500 057, INDIA.
[2]School of Computer & Information Sciences, University of Hyderabad,
Hyderabad – 500 046 INDIA
krishna.gutha@gmail.com , padmarav@gmail.com

## Abstract

In management, business, economics, science, engineering, and research domains, Large Scale Global Optimization (LSGO) plays a predominant and vital role. Though LSGO is applied in many of the application domains, it is a very troublesome and a perverse task. The Congress on Evolutionary Computation (CEC) began an LSGO competition to come up with algorithms with a bunch of standard benchmark unconstrained LSGO functions. Therefore, in this paper, we propose a hybrid meta-heuristic algorithm, which combines an Improved and Modified Harmony Search (IMHS), along with a Modified Differential Evolution (MDE) with an alternate selection strategy. Harmony Search (HS) does the job of exploration and exploitation, and Differential Evolution does the job of giving a perturbation to the exploration of IMHS, as harmony search suffers from being stuck at the basin of local optimal. To judge the performance of the suggested algorithm, we compare the proposed algorithm with ten excellent meta-heuristic algorithms on fifteen LSGO benchmark functions, which have 1000 continuous decision variables, of the CEC 2013 LSGO special session. The experimental results consistently show that our proposed hybrid meta-heuristic performs statistically on par with some algorithms in a few problems, while it turned out to be the best in a couple of problems.

## 1. Introduction

Optimization consists of minimizing or maximizing a real output objective function for real input decision variables within the specified bounds and may or may not include constraints [1,2]. Optimization without constraints is termed unconstrained optimization and optimization with constraints is termed constrained optimization. Optimization has many sub areas which may include multiple objectives, i.e., multi-objective optimization, or which may have multiple good solutions for the same objective function, i.e., multimodal optimization. Optimization has a wide range of applications in mechanics, economics, finance, electrical engineering, operational research, control engineering, geophysics, molecular modeling, etc. Optimization methods are classified into three sub-categories, namely 1) Classical optimization, 2) Heuristic-based optimization and 3) Metaheuristic-based optimization. Classical optimization techniques applied only to convex, continuous and differential search problems, but heuristic and metaheuristic-based optimization can also be applied to non-convex, discontinuous and non-differential search problems. Heuristic-based optimization has some inherent assumption which is problem specific, but metaheuristic-based optimization has no problem-specific assumption [3].

---

[*] Corresponding Author, Phone: +914023294042; FAX: +914023535157

Metaheuristic-based optimization techniques are classified into Evolutionary Computing (EC), Swarm Intelligence-based optimization (SI), Stochastic-based optimization, Physics-based Optimization, Artificial Immune System (AIS)-based optimization, etc. Metaheuristic algorithms are further classified into the population, or point-based, optimization algorithms. Evolutionary computing techniques include the Genetic Algorithm [4], Genetic Programming [5], and Differential Evolution (DE) [6], etc., which are all single objective optimization algorithms. However, Multi-Objective Genetic Algorithm [7], Strength Pareto Evolutionary Algorithm-2 [8], Nondominated Sorting Genetic Algorithm-II (NSGA-II) [9], etc. are examples of multi-objective evolutionary computing algorithms. Swarm Intelligence-based optimization algorithms include Particle Swarm Optimization (PSO) [10], Ant Colony Optimization (ACO) [11], and Bee Swarm-based Optimization [12], etc. Stochastic-based optimization algorithms include Tabu Search [13,14], Stochastic Hill Climbing [15], and Threshold Accepting (TA) [16], etc. Physics-based optimizations include Simulated Annealing [17], Harmony Search (HS) [18], etc. The Artificial Immune System [19,20] based optimization, includes Negative Selection, Clonal Selection, etc.

There is also another class of meta-heuristics that combine the power of more than one meta-heuristic. These are called Memetic Algorithms (MA). MAs have evolved over three generations. First generation MAs are the hybrid optimization algorithms. These use the power of one class of metaheuristic to carry out the exploration and the power of another class of metaheuristic to do a local search. In a survey, presented by Ong and Chen (2011) [21] multi-facets of memetic algorithms are discussed. In [21], the past, present, and future of memetic algorithms is discussed. In this work; we are proposing a first-generation hybrid-iterative two-stage memetic algorithm by combining Harmony Search and Differential Evolution. Harmony Search does the job of exploration and exploitation, and Differential Evolution does the job of giving a perturbation to the HS for coming out of basin (local optima) near optimal value. But here, we are utilizing an Improved Modified Harmony Search (IMHS) proposed by [22] for finding optimal values in reliability complex systems problems. Also, Modified Differential Evolution (MDE) with alternate selection strategy proposed by [23] to detect outliers using evolutionary computing.

In section 2 a literature survey is presented, and in section 3, the motivation for the proposed algorithm is described. In section 4, the list of 15 benchmarks used in the optimization is listed out. In section 5, the basic algorithms used for the construction of the hybrid meta-heuristic are presented. The proposed algorithm with the parameter setting that is utilized is described in section 6. In section 7, results along with statistical analysis and ranking, are presented. Finally, in section 8, a conclusion is made.

## 2. Literature Survey

In Ravi et al. (1997) [24], a novel Improved Non-equilibrium Simulated Annealing (I-NESA) algorithm has been proposed to solve reliability optimization of complex systems with constraints. In Ravi et al. (2000) [25], fuzzy global optimization used to solve complex system reliability is proposed with the incorporation of fuzzy logic in optimization. In Bhat et al. (2006) [26], improved DE is proposed in combination with the reflection property of the simplex method for the efficient parameter estimation of biofilter modeling. In Chauhan and Ravi (2010) [27], a hybrid, based on DE and TA named DETA, is developed for the unconstrained problems and is compared with DE. The results regarding function evaluations show that DETA outperforms DE. Choudhuri and Ravi (2010) [28], proposed a hybrid combining Modified Harmony Search (MHS) and the Modified Great Deluge Algorithm (MGDA) for the unconstrained problems and compared them with MHS. The results prove that MHS+MGDA perform better than MHS. In Maheshkumar and Ravi (2011) [29], a modified harmony search and threshold accepting hybrid was proposed and compared with HS and Modified HS on the unconstrained problems. In Maheshkumar et al. (2013) [30], a hybrid combining both PSO and TA, called PSOTA was developed for unconstrained optimization problems, which gave better optimization results than PSO. Finally, a hybrid optimization algorithm combining ACO, and the classical optimization algorithm called Nelder-Mead simplex was developed to train a neural network for bankruptcy prediction [31].

Large Scale Global Optimization (LSGO) benchmarks were proposed in a special session of Congress on Evolutionary Computation (CEC) 2008 [32]. Later these benchmarks were upgraded in CEC 2010 LSGO special session, and the same was utilized in CEC 2012 LSGO special session [33]. Some of the excellent LSGO algorithms applied on 20 benchmarks LSGO functions were presented in CEC 2010 LSGO special session [34–40]. Another set of LSGO algorithms applied on the same LSGO functions was presented in CEC 2012 LSGO special session [41–45].

In CEC 2010 LSGO special session, MA-SW-Chains of Molina et al.'s (2010) algorithm became the winner of the LSGO competition. In Molina et al. (2010) [35], the MA-SW-Chains algorithm, by chaining different local search applications, assigns to every individual a local search intensity that depends on its features. In Wang and Li (2010) [39], a two-stage-based ensemble optimization evolutionary algorithm (EOEA) is designed to handle LSGO problems. In Omidvar et al. (2010) [37], a systematic way (DECC-DML) is proposed to capture interacting variables for more effective problem decomposition suitable for cooperative coevolutionary frameworks to solve LSGO problems. In Zhao et al. (2010) [40], the dynamic multi-swarm particle swarm optimizer (DMS-PSO) and a sub-regional harmony search (SHS) are hybridized to obtain DMS-PSO-SHS. In DMS-PSO-SHS, sub-swarms are dynamic and small, which is also appropriate to be the population of the harmony search. In Brest et al. (2010) [34], authors present the self-adaptive differential evolution algorithm jDElsgo on large-scale global optimization. In Korosec et al. (2010) [36], Differential Ant-Stigmergy Algorithm (DASA) is developed for LSGO, which transforms a real-parameter optimization problem into a graph-search problem, and then the parameters' differences assigned to the graph vertices are used to navigate through the search space. In Wang et al. (2010) [38], sequential Differential Evolution (DE), enhanced by neighborhood search (SDENS) is proposed, which consists of two steps. In the first step, for every individual, we create two trial individuals by local and global neighborhood search strategies. Second, we select the fittest one among the current individual and the two created trial individuals as a new current individual. The above algorithms in this paragraph are all presented in CEC 2010 LSGO special session.

The Multiple Offspring Sampling (MOS) algorithm became the winner of the CEC 2012 LSGO competition [44]. The Multiple Offspring Sampling (MOS) framework has been used to combine two different heuristics: the first one of the local searches of the Multiple Trajectory Search (MTS) algorithm, and the well-known Solis and Wets heuristic. In Brest et al. (2012) [41], the authors present the self-adaptive differential evolution algorithm (jDEsps) with a small and varying population size on large-scale global optimization. In Takahama and Sakai (2012) [45], the performance of Differential Evolution with Landscape Modality Detection and a Diversity Archive (LMDEa) reported on the set of benchmark functions provided for the CEC 2012 Special Session on LSGO using a small population size and a large archive for diversity. In Fister et al. (2012) [42], a Memetic ABC (MABC) algorithm was developed and hybridized with two local search heuristics: the Nelder-Mead algorithm (NMA) and the Random Walk with Direction Exploitation (RWDE). The former tends more towards exploration, while the latter more towards exploitation of the search space. In Zhang and Li (2012) [43], a Cooperative Coevolution Evolutionary Algorithm (CCEA) with Global Search (CCGS) is presented to handle the premature convergence of CCEA and applied to LSGO problems.

Recently, two new LSGO algorithms, namely the Memetic Algorithm with Adaptive Local Search Depth (MA-ALSD) in Liu and Li (2014) [46] and the Joint Operations Algorithm (JOA) in Sun et al. (2016) [47] have been proposed with CEC 2010 LSGO benchmarks. The MA-ALSD presented in CEC 2014 was with the CEC 2010 LSGO benchmarks. In Sun et al. (2016) [47], JOA was compared with six other metaheuristics. First, two variants of DE, namely dynamic Group-based Differential Evolution (GDE) and Sinusoidal Differential Evolution (SinDE) were compared. The next two variants compared are the particle swarm optimizer, with a diversity enhancing mechanism & neighborhood search strategies (DNSPSO) ,and a dynamic multi-swarm particle swarm optimizer, with a cooperative learning strategy (D-PSO-C). Lastly, another two novel algorithms, namely, Free Search (FS) and Social Based Algorithm (SBA) are compared. Of all the algorithms that were compared, JOA gave the best results with regards to the mean optimal value.

Mahdavi et al. (2015) [48] surveyed the LSGO problems, with a good classification of LSGO algorithms that is based on the approach used for large scale optimization. A description of the application of LSGO algorithms used for the various real-world problems, like data mining, mechanics, etc., is also surveyed in the above article. In LaTorre et al. (2015) [49], a comprehensive comparison of all the winning LSGO algorithms in various LSGO competitions, was made. A comparison of the algorithms of all the LSGO winners (along with other competition winners) is also made on the CEC 2013 LSGO benchmarks.

## 3. Motivation

Das et al. (2011) [50] show that the explorative power of HS is initially is very good, but that it falls, as the number of iterations increase, and it gets stuck at the local optimal basin, without progressing further. To overcome this, HS is to be perturbated with a good and powerful meta-heuristic. In Kim et al. (2016) [51], various performance measures such as dimensionality, the number of local optima, interval span of side constraints, the ratio of local optima, valley structure coefficient, and peak density ratio were compared to a set of good meta-heuristics. These compared meta-heuristics are as follows: Random Search (RS), Simulated Annealing (SA), Particle Swarm Optimization (PSO), Water Cycle Algorithm (WCA), Genetic Algorithms (GAs), Differential Evolution (DE), Harmony Search (HS), and Cuckoo Search (CS). The compared DE gave better results than all those with a good coverage, and HS gave comparatively better results than some with second-best coverage when a radar plot was created with various performance measures. This gave us the motivation to combine HS with DE to obtain a better optimization algorithm as DE gave better results for all performance measures.

We combined IMHS, which was developed for complex reliability systems in [22], with differential evolution. We also developed a modified DE with an alternate selection strategy in [23]. This MDE does the exploitation well as the selection of DE is replaced with that of HS where a better solution replaces the worst solution instead of the current solution vector. Selection operation usually suggests the exploitation power of any evolutionary algorithm as suggested in Das et al. (2011) [50]. So, we replaced the DE selection strategy with that of HS, as it replaces the worst solution vector, which is better than DE regarding exploitation.

Also, we tested our hybrid initially with PSOTA, on the 30-dimensional problems [30]. Our hybrid gave better results with regard to the success rate and function evaluations, as compared to that of PSOTA, which gave us the initial fuel to test on LSGO problems of CEC 2013 special session. Other criteria for choosing the concerned algorithms for hybridization are that they have fewer parameters to tune. IMHS has dynamic tuning of one parameter while another parameter is fixed and MDE has two parameters to tune. Thus, a total of two parameters are needed to be tuned in the hybrid.

## 4. CEC 2013 LSGO Benchmarks

A large number of meta-heuristics are developed in the recent times. For all these meta-heuristics, performance gradually decreases as dimensionality increases. This problem with high dimensionality is termed "Curse of Dimensionality" [60]. Also, many real world problems are high dimensional in nature. The following describes why high dimensional problems are hard to solve:

- Search space explodes as dimensionality increases
- High dimensional problem evaluation takes a lot of resources and is usually costly
- There may be a dependency between variables
- Search space properties may change as dimensionality increases.

To address these issues of high dimensional problems, CEC, in 2008, started a special session to come up with large-scale global optimizing algorithms. This initial LSGO special session uses only simple functions in their test suite. Later, in CEC 2010 and CEC 2012, it has been extended to make this special session a suitable evaluation platform for testing and comparing large-scale global optimization. CEC 2013 LSGO special session [60] was developed by keeping in mind the following four types of high-dimensional problems: 1) separable functions, 2) partially-separable functions, 3) functions that consist of overlapping subcomponents and 4) fully-nonseparable functions. In CEC 2013 LSGO special session are developed by adding the following features to the high dimensional problems :

- Nonuniform subcomponent sizes;
- Imbalance in the contribution of subcomponents;
- Functions with overlapping subcomponents;
- New transformations to the base functions:
    – Ill-conditioning;
    – Symmetry breaking;
    – Irregularities.

The basic functions used in the CEC 2013 LSGO special session are:
1. The Sphere Function
2. The Elliptic Function
3. Rastrigin's Function
4. Ackley's Function.
5. Schwefel's Problem 1.2
6. Rosenbrock's Function

To these basic functions, 1000 dimensional input vectors are passed, which are computed from the optimization algorithms. These input vectors are shifted, permuted, and rotated (rotated by multiplying with the orthogonal matrix). By applying these shifting, permuting, and rotation operations, the job of the optimization algorithm is made difficult [60].

Of these basic functions, some are unimodal and others are multimodal. Sphere, Elliptic and Schwefel's functions are unimodal, as they have unique local optima. Rosenbrock's, Rastringin's and Ackley's functions are multimodal, as they have multiple local optima. Some are separable, partially separable and fully non-separable. Sphere, Elliptic, Rastringin's and Ackley's functions are separable. Some of these are made partially separable by multiplying some parts of the input vector with the orthogonal matrix. However, Schwefel's and Rosenbrock's functions are fully non-separable. Some of the functions are scalable, which means that they can be applied to problems with a dimenision of more than 1000.

By considering the above factors 15 benchmarks have been designed. All of these benchmark functions are unconstrained. These 15 benchmark functions of the CEC 2013 LSGO special session, with all the above factors, are summarized in Table 1.

Table 1. Summary of functions used in CEC 2013 LSGO competition

| | a. Fully-separable Functions |
|---|---|
| 1 | F1: Elliptic Function |
| 2 | F2: Rastrigin Function |
| 3 | F3: Ackley Function |
| | b. Partially Additively Separable Functions |
| | i) Functions with a separable subcomponent: |
| 4 | F4: Elliptic Function |
| 5 | F5: Rastrigin Function |
| 6 | F6: Ackley Function |
| 7 | F7: Schwefels Problem 1.2 |
| | ii) Functions with no separable subcomponents: |
| 8 | F8: Elliptic Function |
| 9 | F9: Rastrigin Function |
| 10 | F10: Ackley Function |
| 11 | F11: Schwefels Problem 1.2 |
| | c. Overlapping Functions |
| 12 | F12: Rosenbrock's Function |
| 13 | F13: Schwefels Function with Conforming Overlapping Subcomponents |
| 14 | F14: Schwefels Function with Conflicting Overlapping Subcomponents |
| | d. Non-separable Functions |
| 15 | F15: Schwefels Problem 1.2 |

## 5. Optimization Algorithms Utilized

### 5.1 Nature of Optimization Problem

The current optimization problem has 15 objective functions of the form F: R → R. When given input from a set of bounded real number domains, it gives an output in the real number range. All 15 objective functions are minimization problems by nature. For minimization problems, the solution vector generated from the optimization algorithm, '$X_{new}$', when given to the function, 'F', should be less than '$X_{old}$', i.e., $F(X_{new}) \leq F(X_{old})$.

## 5.2 Improved Modified Harmony Search

Harmony Search (HS) is a process mimicking metaheuristic proposed by Geem et al. (2001) [18]. It mimics the musician's improvisation process of playing an old piece of music or generating new music from the piece of old music or generating a new music from scratch.

Three parameters used in Harmony Search namely are Harmony Memory Size (HMS), Harmony Memory Considering Rate (HMCR), and Pitch Adjusting Rate (PAR). Discussed below are the parameters of HS:

- HMS = the size of the harmony memory. It can vary from 30 to 300.
- HMCR = the rate of choosing a value from the harmony memory. (1-HMCR) is used to control exploration
- PAR = the rate of choosing a neighboring value. It is used to control exploitation

Harmony update is an important part of HS, which generates new solution vectors. Below are the steps that describe Harmony Update:

1. Pick a solution from harmony memory if the random number, generated between 0 and 1, is less than the HMCR.
    i) Search in the local neighborhood of the picked solution vector if the random number, generated between 0 and 1, is less than PAR.
    ii) If the random number generated between 0 and 1 greater than the PAR, use the same picked solution vector.
2. If the random number generated between 0 and 1 greater than the HMCR, generate random values in between the bounds of the input decision variable.

The general steps for Harmony Search are described below:

1. Firstly, fix the size of harmony memory (HM).
2. Initialize the HM number of solution vectors within bounds randomly and compute corresponding fitness values for the solution vectors.
3. HMCR is linearly increased between 0.7 and 0.9, and the PAR fixed at a constant is between 0 and 1.0.

4. The harmony update is performed i.e. using old solution vectors or generating a new solution vector from the old solution vector or generating a totally new solution vector utilizing the concerned HMCR and PAR values.
5. If the fitness of the new solution vector generated from the harmony update is better than the worst solution vector in the harmony memory, then replace the worst solution vector with the new solution vector. Also, replace the corresponding fitness value of the worst solution vector with the new one.
6. Repeat steps 3, 4, and 5 until the maximum number of iterations.

## 5.3 Modified Differential Evolution

Differential Evolution (DE) is a stochastic population-based evolutionary metaheuristic [6]. DE is developed for real input parameter based optimization problems. It includes four important sections, which are mainly initialization, mutation, recombination and selection. At every iteration, new solutions are created with the combination of solutions chosen from the present set of solutions are called 'mutations'. The solutions that are obtained at this stage are combined with a specific target solution in a process called 'crossover'. At last, based on fitness, the solutions are selected. The following steps briefly describe the working of DE, which are also illustrated in Fig. 1:

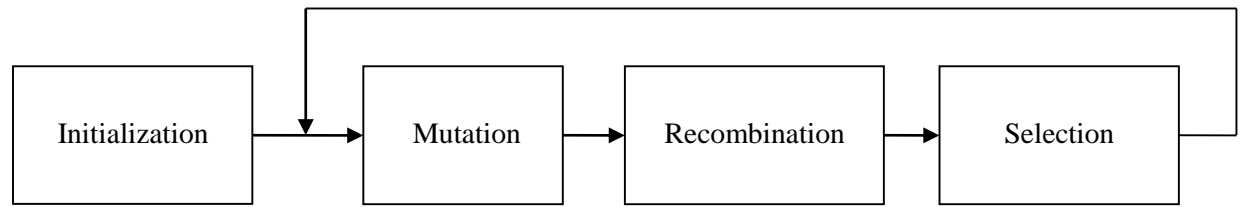

Figure 1: Initialization in DE algorithm

**Step-1: Initialization**

1. Initialize *NP*, the number solution vectors of dimensionality size, *Dim*.

2. Check whether the solution vectors are within bounds.

3. Compute the fitness values of the NP number of solution vectors.

**Step-2: Mutation**

4. Let $x_1$, $x_2$, $x_3$ and $x^{current}$ be the three unique solution vectors chosen randomly from the population *NP*.

5. Perform the mutation using the following equation

$$u'_i = x_{1,i} + F * (x_{2,i} - x_{3,i})$$

The above equations are two different variants of DE. First one is of *DE/rand/1/bin* and second one is of *DE/best/1/bin* (We present both as we have used both the variants of DE in our proposed method). Differential weight, '*F*', is between 0 and 2.

**Step-3: Recombination**

6. Using the mutation vector and crossover rate parameter *CR* recombination is performed as follows:

$$v_i = \begin{cases} u_i & if(rand(0,1) \leq CR \mid i = I_{rand}) \\ x_i^{current} & if(rand(0,1) > CR \text{ and } i \neq I_{rand}) \end{cases}$$

*CR* is between 0 and 1. $I_{rand}$ is a random dimension between 1 to *Dim* and $I_{rand}$ is used so that *v* and $x^{current}$ are different.

**Step-4: Selection**

7. If Fitness(*v*) ≤ Fitness($x^{worst}$) replace current vector with solution vector *v* in the population.

8. Else do not replace the current vector.

9. Repeat steps 4-8, until a maximum number of iterations.

DE has three parameters: Population Size (NP), Differential Weight (F) and Crossover Rate (CR). Differential weight acts as the step-size of how the DE algorithm moves in the search space. If it is high, it may miss several optima and, if it is small, the convergence toward optima is slow. So, a careful experimentation for the fixation of the F is needed. Another parameter is crossover rate, which controls the recombination rate of the algorithm. If CR is high, more exploration performed and, if low, more exploitation performed.

## 6. Proposed Hybrid Optimization Algorithm

In the proposed hybrid optimization algorithm, the Improved Modified Harmony Search (IMHS) is proposed in [22], which has an 'hmcr' value linearly increasing from 0.7 to 0.9. This is taken and is coupled with Modified Differential Evolution (MDE), proposed in [23], where the selection strategy of DE is replaced with that of HS and where the worst solution will be replaced instead of current solution. The reason for using MDE is perturbating the solutions generated by IMHS as the IMHS has the limitaion of getting stuck at the local optima. Also, Kim et al. (2016) [51] show that HS and DE are the powerful metaheuristics when compared with various performance measures. In Das et al. (2011) [50], it is shown that the power of HS depends on the selection strategy of HS, and also, the limitation of HS is that it gets stuck at the local optima. So, these readings gave us an intuition to combine IMHS with MDE. In the proposed hybrid the population of solution vectors from IMHS is passed to MDE as shown in Fig. 2 iteratively in a loop.

The proposed hybrid optimization algorithm works in two phases cyclically. In the first phase, IMHS is run for a fixed number of generations. The resulting population from IMHS is then passed to MDE in the second phase, where MDE is run for a fixed number of generations. The population from MDE is then passed on to IMHS. Both these phases are repeated (cycled) till the maximum number of iterations is reached. This population is swapped after a fixed number of generations between IMHS to MDE and also from MDE to IMHS till the maximum outer iterations. At first, IMHS, is run for 10000 generations, and then MDE is run for 20000 generations. So, a total of (10000 (IMHS) + 20000 (MDE)) = 30000 generations. One pass through IMHS and MDE is termed one outer iteration. The algorithm is run for 100 passes meaning that a total of (100*30,000) = 3,000,000 inner iterations or FE's take place.

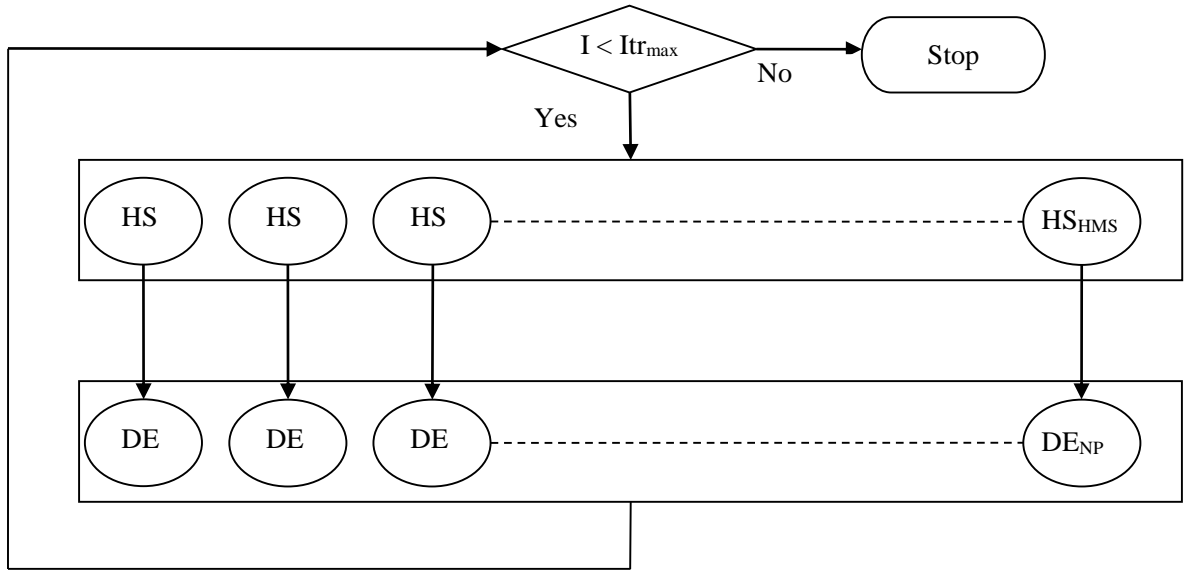

Figure 2: The Process of coupling between HS memory and DE population vectors

**PHASE I: IMHS**

The design flow of IMHS is depicted in Fig. 3.

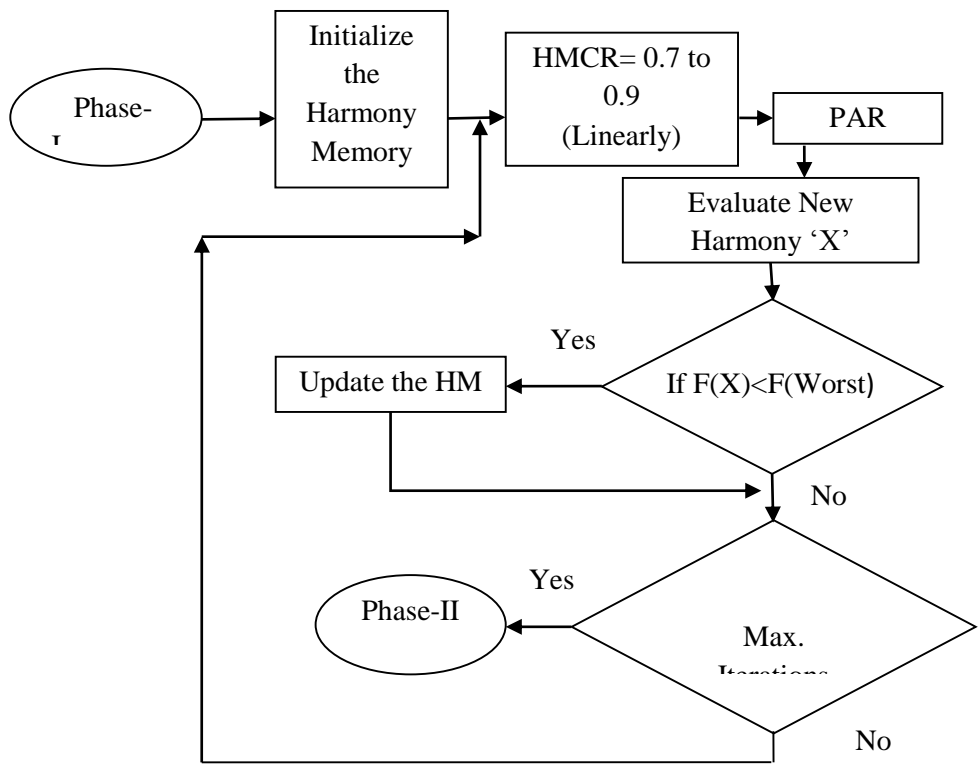

Figure 3: IMHS Design Flow

## PHASE I: MDE

The design flow of MDE is depicted in Fig. 4.

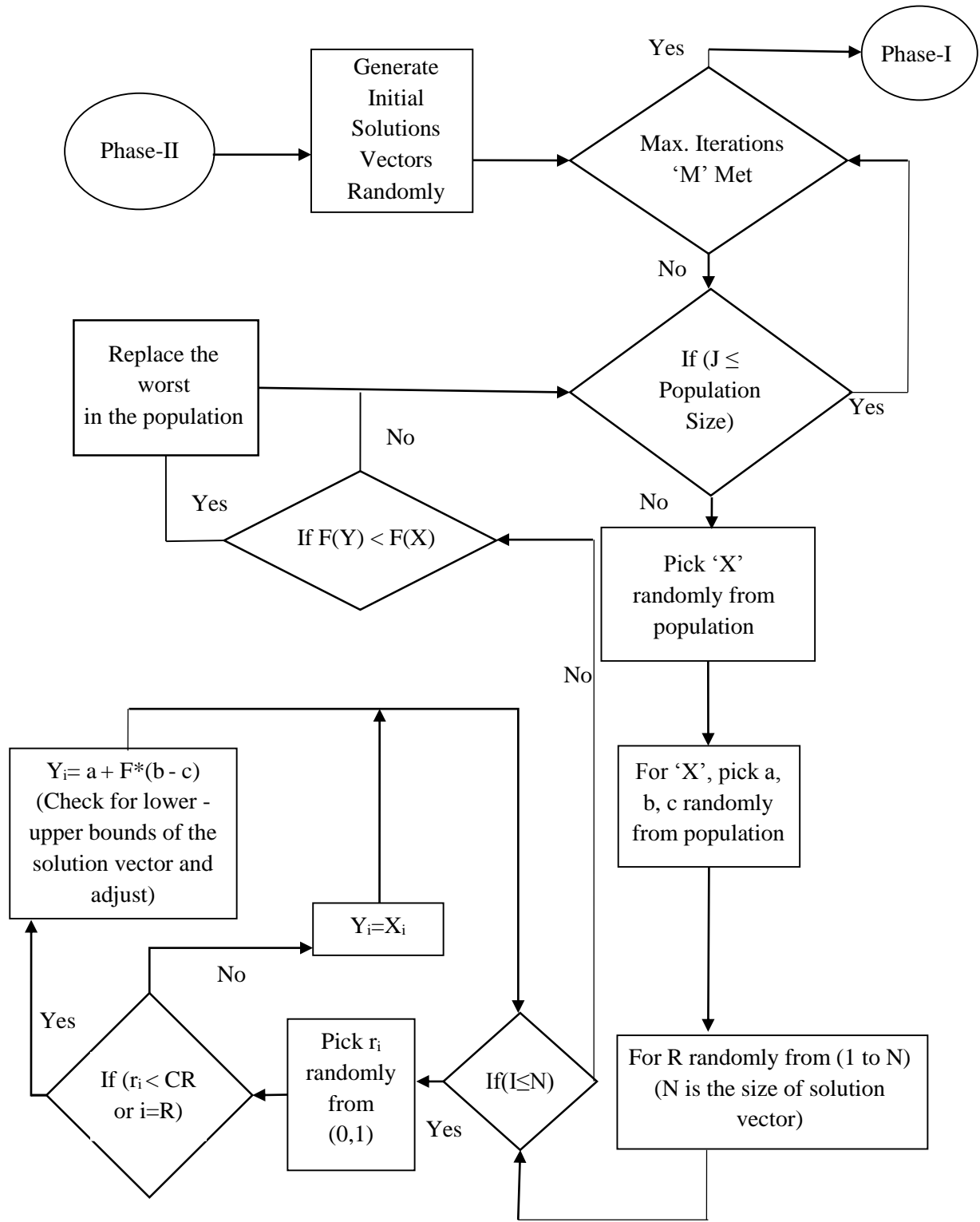

Figure 4: MDE Design Flow

Figures 5-7 present the algorithms for the proposed hybrid large scale optimization algorithm

```
Algorithm 3: IMHS+MDE
Input: PopulationSize = 200, MaxIterations_{IMHS} = 10000, MaxIterations_{MDE} = 100,
       MaxIterations_{Total} = 100
Compute for once Fitness 'Y' of Randomly Initialized Population Vectors 'P'
P_{HM} = P
Y_1 = Y
while K ≤ MaxIterations_{Total} do
    Reset 'HMCR' and MaxIterations Counters of IMHS and MDE
    (P_{DE}, Y_2) = IM-HarmonySearch(P_{HM}, Y_1)
    (P_{HS}, Y_1) = M-DifferentialEvolution(P_{DE}, Y_2)
    if K = 4 | K = 20 | K = MaxIterations_{Total} then
        P = P_{HS}
        Y = Y_1
        Index = IndexofMin(Y)
        Print(P[Index], Y[Index])
    end
    K = K + 1
end
```

Figure 5: Algorithm of Hybrid IMHS+MDE

```
Algorithm 1: Improved and Modified Harmony Search (IMHS)
Input: Initialized Harmony Memory (HM), HMCR and PAR
I = 1
IM-HarmonySearch(HM, Y)
{
    while I ≤ MaxIterations_{IMHS} do
        Increment HMCR from 0.7 to 0.9 in steps
        x' = HarmonyUpdate(HM, hmcr)
        y' = FitnessFx(x')
        if y' better than worst solution vector in HM then
            Replace worst vector by x'
            Also replace the corresponding fitness value with y'
        end
        I = I + 1
    return (HM, Y)
end
}
HarmonyUpdate(HM, hmcr)
{
if rand(0, 1) < hmcr then
    Randomly pick a solution vector from the harmony memory
    if rand(0, 1) < par then
        Modify the solution vector values
        (We have to note that 'par' value is 0.4)
    end
else
    Generate a new random solution vector
end
Check whether the solution vector is within Bounds
}
```

Figure 6: Algorithm of Improved and Modified Harmony Search

```
Algorithm 2: Modified Differential Evolution (MDE)
Input: Population (P_DE), NP, F and C
I = 1
J = 1
M-DifferentialEvolution(P_DE, Y)
{
while I ≤ MaxIterations_MDE do
    while J ≤ NP do
        int x = floor(rand(1, NP + 1))
        int a, b, c
        repeat
        |   a = floor(rand(1, NP + 1))
        until a = x;
        repeat
        |   b = floor(rand(1, NP + 1))
        until b = x | b = a;
        repeat
        |   c = floor(rand(1, NP + 1))
        until c = x | c = a | c = b;
        R = floor(rand(1, Dim + 1))
        Ind1 = P_DE[a]
        Ind2 = P_DE[b]
        Ind3 = P_DE[c]
        Current = P_DE[x]
        Candidate = P_DE[x]
        while K ≤ Dim do
            if floor(rand(1, Dim + 1)) = R or rand(0,1) < CR then
            |   ck = Ind1[K] + F * (Ind2[K] - Ind3[K])
            else
            |   ck = Current[K]
            end
            if ck with in bounds then
            |   Candidate[K] = ck
            |   K = K + 1
            end
        end
        K = 1
        x' = Candidate
        y' = FitnessFx(x')
        if y' better than worst solution vector in P_DE then
            Replace worst vector by x'
            Also replace the corresponding fitness value with y'
        end
        J = J + 1
    end
    I = I + 1
end
return (P_DE, Y)
}
```

Figure 7: Algorithm of Modified Differential Evolution

In the proposed hybrid optimization, the modified version of HS is coupled with the modified version of DE. The proposed hybrid has the following modifications to the basic algorithm of HS and DE:

- Increment HMCR linearly from 0.7 to 0.9 as HS performs better within this range (Yang, 2009). This modification is needed for good scope for exploration when HMCR is at 0.7 and when HMCR reaches 0.9, exploitation is performed with slight scope for exploration. In the hybrid MHS+TA discussed in the literature survey above, experimentation performed with the HMCR value is incremented linearly from 0 to 1. Though the hybrid MHS+TA gave good results, the convergence was slow as a lot of exploration is performed, which is unnecessary. And, also, when the HMCR value reaches one (1) no more exploration is performed, which causes the algorithm to get stuck at local optima. So, we came up with a better version of HS, where the HMCR is incremented from 0.7 linearly up to 0.9.
- PAR value is fixed by parameter tuning using 'GAReal' function of 'gaoptim' package in R. The PAR helps in local improvements, and some of the previous solution vector values are retained without getting a totally new vector out of exploitation.
- The population size is another parameter which has to be carefully chosen as hybrid yields better results only when the coupling is performed well with right population sizes. Various experimentations with different population sizes i.e. 30, 200, and 300, was performed for a fixed number of function evaluations and 1000 dimensions. Initially, the coupling between population sizes, 30, and 200, yielded incrementally better results. But, after 200, the coupling yielded a decrease in performance as well as no improvement (meaning a lot of overhead for CPU without much improvement in the betterment of the Optima). Also, the reason for choosing 200 as population size is that DE will have a better scope of the alternate selection strategy (discussed below) for more number of diverse solution vectors in the population than less number of solutions.
- As discussed in the motivation section, the HS algorithm's selection strategy is utilized in DE making it MDE. The reason is that HS has better exploitative power, with worse vector replacement than DE's current vector replacement. By doing so, if the current vector is better than worst, it remains in the population, rather than getting replaced. By doing so, the next time the DE will have a chance to search the local neighborhood near the current vector.
- Parameters F and CR are set by careful parameter tuning using 'GAReal' function of 'gaoptim' package in R. Choosing a low value of F gives a chance to search slowly and finely. If CR is set not too high or too low, then it helps in finding the right mix between mutation and recombination by searching the local neighborhood better.
- The maximum number of inner iterations for the MDE is set to double those of IMHS as more exploitation is needed than exploration. The above helps in fine-tuning to the solution, when more exploitation is performed rather than skipping the solution. Another reason for setting MDE iterations to double those of IMHS is that the step size of MDE is kept low so that it can do fine the tuning well. Also, this cycle of first performing IMHS and then performing MDE is repeated as one contributes to the diversity of the population of the other by repeating iteratively. We have also experimented with different settings of inner iterations to set for IMHS and MDE, and the current setting of inner iterations gave better results with faster convergence.

- The coupling happening between IMHS and MDE is a form of loose coupling, as the optimization algorithms can be separated from each other, whereas in the tight coupling the optimization algorithms cannot be separated which is not the case here in the proposed approach.

## 6.1 Time complexity of the proposed hybrid

The approach proposed in Das and Suganthan (2011) [53] for DE to compute the time complexity is employed for both the algorithms. The algorithm runs for a total of three million function evaluations (FE's). IMHS, at first, runs for 10000 FE's, and MDE, next, runs for (100*200) = 20000 FE's where 100 represents inner iterations and 200 represents the population size. So, a total of (10000 (IMHS) + 20000 (MDE)) = 30000 FE's (this is one outer iteration). These outer iterations repeated 100 times (for a total of (100*30,000) = 3,000,000 FE's).

The time complexity for IMHS is (number of inner iterations ($I_{IMHS}$) * number of outer iterations (OI) * dimensionality of the objective function (D)) = 10000*100*D = 1,000,000*D = 1/3$^{rd}$ of Total FE's (TFE$_{IMHS}$) * D. So, the time complexity is represented as O(TFE$_{IMHS}$*D). The time complexity for MDE is (number of inner iterations ($I_{MDE}$) * number of outer iterations (OI) * Population Size (PS) * dimensionality of the objective function (D)) = 100*100*200*D = 2,000,000*D = 2/3$^{rd}$ of Total FE's (TFE$_{MDE}$) * D. So, the time complexity is represented as O(TFE$_{MDE}$*D). Combining both we get O((TFE$_{IMHS}$ + TFE$_{MDE}$)*D).

## 6.2 Parameter Tuning for the proposed hybrid

### 6.2.1. Specialist Parameter Tuning Methodology

There has been much debate on using a generalist vs. a specialist way of parameter tuning for Evolutionary Algorithms (EA) [61, 62]. In the current work, we employed a specialist way of parameter tuning. The reason is that according to No Free Lunch Theorem [63] there cannot be an EA, which is uniformly the best for all the optimization problems. As the decision space changes, the parameters used for the EA for the decision space also change. Therefore, a generalist way of parameter tuning is not suitable here. Hence, we employed a specialist way of parameter tuning.

Further, according to Eiben and Smit [62]

"Tuning an EA on a set of functions delivers a generalist, that is, an EA that is good at solving various problem instances. Obviously, a true generalist would perform well on all possible functions. However, this is impossible by the no-free-lunch theorem. Therefore, the quest for generalist EAs is practically limited to less general claims that still raise serious methodology issues"

Due to the afore-mentioned serious issues, as quoted in the above quoted paragraph, we choose to adopt the specialist methodology for the parameter tuning of the non-adaptive parameters. As it can be seen, most of the parameters of the EAs used for most LSGO algorithms are adaptive (i.e. no need of parameter tuning). But, our proposed hybrid EA is a mix of adaptive and non-adaptive parameters.

**6.2.2. Parameter List and Tuning Methodology**

The parameter tuning for the proposed hybrid method for each CEC 2013 LSGO function is performed by running a stochastic optimization on top of the hybrid method. We ran the function called 'GAReal' of 'gaoptim' package of 'R' on top of the proposed hybrid for a fixed number of iterations. Three parameters namely PAR, CR, and F are fine-tuned by the above R function (GAReal). The bounds for the parameters PAR and CR are between 0 and 1. The bounds for the parameter F is between 0 and 2. The HMCR is not tuned as it is linearly changed between 0.7 and 0.9. Tables 2a and 2b presented, shows the different parameter values of the proposed hybrid for the 15 benchmark functions of the CEC 2013 LSGO.

Table 2a: Parameters for the IMHS+MDE

| Parameters | Values and Ranges |
|---|---|
| **Population Size** | 200 |
| **HMCR** | 0.7 to 0.9 (Linear Increment) |
| **PAR** | 0.3 to 0.6 (Non-adaptive) (Fixed Values in the Range by Specialist Methodology - See Table 2b) |
| **CR** | 0.15 to 0.95 (Non-adaptive) (Fixed Values in the Range by Specialist Methodology- See Table 2b) |
| **F** | 0.3 to 0.7 (Non-adaptive) (Fixed Values in the Range by Specialist Methodology - See Table 2b) |

Table 2b: Specialist Method of Parameter tuning for PAR, CR, and F

|  | F1 | F2 | F3 | F4 | F5 | F6 | F7 | F8 | F9 | F10 | F11 | F12 | F13 | F14 | F15 |
|---|---|---|---|---|---|---|---|---|---|---|---|---|---|---|---|
| **PAR** | 0.3 | 0.74 | 0.4 | 0.34 | 0.10 | 0.56 | 0.42 | 0.34 | 0.37 | 0.15 | 0.49 | 0.45 | 0.39 | 0.41 | 0.34 |
| **CR** | 0.36 | 0.42 | 0.78 | 0.94 | 0.40 | 0.83 | 0.79 | 0.94 | 0.15 | 0.20 | 0.80 | 0.77 | 0.76 | 0.79 | 0.14 |
| **F** | 0.3 | 0.3 | 0.51 | 0.47 | 0.29 | 0.69 | 0.48 | 0.39 | 0.37 | 0.56 | 0.46 | 0.49 | 0.51 | 0.48 | 0.51 |

We can visualize the specialist methodology of parameter values present in Table 2b using a box plot depicted in Figure 8, which gives readers an intuitive idea of the distribution of the parameters. Also the box plot indicates that our proposed algorithm is more susceptible to the CR compared to other two because of the higher its variability and therefore tuning the parameter CR is harder.

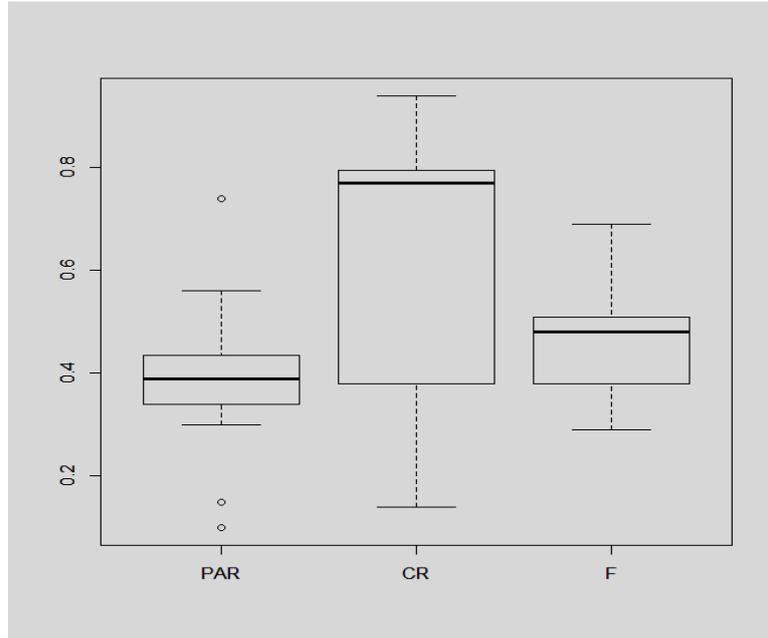

Figure 8: Box Plot of the non-adaptive parameters

Another facet of our proposed algorithm for LSGO is that it is suitable for single objective unconstrained optimization framework. Therefore, the specialist methodology of parameter tuning for each of the problems is apt.

## 7. Results Analysis

The proposed algorithm along with the CEC 2013 LSGO benchmarks are implemented in R. The algorithms are executed on Intel(R) Core(TM) i7-6700 @ 3.4 GHz processor and 32 GB RAM. The Operating System used is Windows 8.1 Pro 64 bit.

The proposed algorithm are compared with the winners of the CEC 2010 LSGO, SOCO 2011, CEC 2012 LSGO and CEC 2013 LSGO competitions on the CEC 2013 LSGO benchmarks. The results for comparison are taken from LaTorre et al. (2015) [49]. The algorithms of winners of the four competitions listed below (along with competition ranks):

1) CEC 2010 LSGO
    i) MA-SW-Chains (Rank 1) [35]
    ii) 2S-Ensemble (Rank 2) [39]
2) SOCO 2011
    i) Multiple Offspring Sampling (MOS) - 2011 (Rank 1) [54]
    ii) jDElscop (Rank 2) [55]
    iii) GaDE (Rank 3) [56]
3) CEC 2012 LSGO
    i) Multiple Offspring Sampling (MOS) - 2012 (Rank 1) [44]

ii) jDEsps (Rank 2) [41]
4) CEC 2013 LSGO
   i) Multiple Offspring Sampling (MOS) - 2013 (Rank 1) [57]
   ii) DECC-G (Rank 2) [58]
   iii) CC-CMA-ES (Rank 3) [59]

The results of the proposed algorithm along with the other compared algorithms are presented in Tables 3-4. The functions, wherein the proposed algorithm beats the other algorithms, are marked in yellow, and the functions where it came second or third are marked in green. The functions for which the compared algorithms performed better than all the other algorithms are also marked in yellow color.

The proposed algorithm is run for fixed FEs as per the CEC 2013 LSGO benchmark (i.e. 3 million FEs for each of the 25 independent runs). Median value of 25 independent runs (i.e. 13th) is used for the statistical analysis as per the benchmark. The winner is decided based on the median value of 25 independent runs (i.e. 13th in the increasing order) for each function.

Table 3: Results of CEC 2010, CEC 2012, CEC 2013 and SOCO 2011 LSGO winners and proposed hybrid algorithms applied on CEC 2013 LSGO Benchmarks (F1, F2, F3, F4, F5, F6, F7)

| Algorithm | Metrics | F1 | F2 | F3 | F4 | F5 | F6 | F7 | F8 |
|---|---|---|---|---|---|---|---|---|---|
| **IMHS+MDE (Proposed)** | Best | 943082.9 | 1454.525 | 20.65726 | 4.64E+09 | 590036.6 | 1058137 | 92875120 | 3.03E+13 |
| | Median | 4595495 | 1564.223 | 20.68594 | 9.75E+09 | 932950.4 | 1059663 | 1.42E+08 | 1.53E+14 |
| | Worst | 13521885 | 1642.277 | 20.70005 | 1.2E+10 | 1193703 | 1061637 | 2.36E+08 | 3.26E+14 |
| | Mean | 4654139 | 1556.606 | 20.6803 | 9.25E+09 | 947524.7 | 1060032 | 1.5E+08 | 1.75E+14 |
| | Stddev | 3127253 | 54.91001 | 0.014517 | 1.96E+09 | 150477.7 | 1049.047 | 41171630 | 8.46E+13 |
| **M A-SW-Chains** | Best | 3.03E-14 | 9.52E+02 | 2.63E-13 | 7.20E+08 | 1.07E+06 | 9.54E-01 | 3.36E+05 | 3.14E+13 |
| | Median | 6.15E-13 | 1.13E+03 | 6.79E-13 | 2.70E+09 | 1.98E+06 | 6.24E+02 | 3.99E+06 | 4.65E+13 |
| | Worst | 4.97E-12 | 1.50E+03 | 1.24E-12 | 9.06E+09 | 7.59E+06 | 5.79E+04 | 4.40E+06 | 6.42E+13 |
| | Mean | 1.14E-12 | 1.18E+03 | 6.78E-13 | 3.80E+09 | 2.26E+06 | 1.07E+04 | 3.78E+06 | 4.63E+13 |
| | Stddev | 1.28E-12 | 1.84E+02 | 2.28E-13 | 2.70E+09 | 1.36E+06 | 2.09E+04 | 8.46E+05 | 9.18E+12 |
| **2S-Ensemble** | Best | 0.00E+00 | 1.04E-26 | 2.81E-13 | 4.18E+09 | 5.89E+05 | 1.59E+05 | 6.88E+05 | 1.21E+14 |
| | Median | 0.00E+00 | 9.95E-01 | 3.98E-13 | 1.13E+10 | 1.28E+06 | 2.00E+05 | 1.54E+06 | 3.94E+14 |
| | Worst | 4.55E-23 | 5.97E+00 | 5.72E-13 | 1.68E+10 | 1.79E+06 | 2.25E+05 | 4.17E+06 | 6.82E+14 |
| | Mean | 2.35E-24 | 1.39E+02 | 4.20E-13 | 1.10E+10 | 1.33E+06 | 1.94E+05 | 1.90E+06 | 3.85E+14 |
| | Stddev | 9.16E-24 | 1.62E+00 | 7.82E-14 | 3.44E+09 | 2.50E+05 | 1.64E+04 | 1.14E+06 | 1.39E+14 |
| **MOS-SOCO2011** | Best | 0.00E+00 | 1.10E+01 | 0.00E+00 | 3.14E+09 | 8.51E+06 | 9.80E+05 | 9.05E+05 | 3.92E+14 |
| | Median | 0.00E+00 | 1.96E+01 | 0.00E+00 | 1.20E+10 | 1.09E+07 | 9.85E+05 | 1.23E+07 | 1.13E+15 |
| | Worst | 0.00E+00 | 2.72E+01 | 0.00E+00 | 3.24E+10 | 1.44E+07 | 9.91E+05 | 2.20E+08 | 8.82E+15 |
| | Mean | 0.00E+00 | 1.93E+01 | 0.00E+00 | 1.34E+10 | 1.11E+07 | 9.85E+05 | 2.31E+07 | 1.64E+15 |
| | Stddev | 0.00E+00 | 4.16E+00 | 0.00E+00 | 7.69E+09 | 1.79E+06 | 3.22E+03 | 4.42E+07 | 1.66E+15 |
| **jDElscop** | Best | 0.00E+00 | 0.00E+00 | 2.56E-13 | 1.80E+09 | 3.24E+06 | 3.00E-07 | 9.32E+08 | 3.51E+13 |
| | Median | 0.00E+00 | 9.95E-01 | 2.77E-13 | 7.18E+09 | 4.83E+06 | 1.99E-01 | 3.17E+09 | 1.05E+14 |
| | Worst | 0.00E+00 | 1.10E+01 | 2.91E-13 | 2.05E+10 | 6.17E+06 | 5.66E+03 | 5.89E+09 | 2.44E+14 |
| | Mean | 0.00E+00 | 1.92E+00 | 2.78E-13 | 8.37E+09 | 4.78E+06 | 2.42E+02 | 3.29E+09 | 1.18E+14 |
| | Stddev | 0.00E+00 | 2.67E+00 | 8.41E-15 | 4.15E+09 | 7.60E+05 | 1.13E+03 | 1.24E+09 | 5.28E+13 |
| **GaDE** | Best | 0.00E+00 | 0.00E+00 | 2.17E-13 | 4.68E+09 | 8.72E+06 | 1.02E+06 | 1.97E+09 | 4.87E+13 |
| | Median | 0.00E+00 | 0.00E+00 | 2.56E-13 | 1.01E+10 | 1.02E+07 | 1.03E+06 | 2.68E+09 | 1.57E+14 |
| | Worst | 0.00E+00 | 1.72E-02 | 3.16E-13 | 2.79E+10 | 1.18E+07 | 1.03E+06 | 3.23E+09 | 8.15E+14 |
| | Mean | 0.00E+00 | 6.88E-04 | 2.61E-13 | 1.12E+10 | 1.02E+07 | 1.03E+06 | 2.67E+09 | 2.08E+14 |
| | Stddev | 0.00E+00 | 3.44E-03 | 2.95E-14 | 5.17E+09 | 9.06E+05 | 2.03E+03 | 3.90E+08 | 1.75E+14 |
| **MOS-CEC2012** | Best | 0.00E+00 | 1.57E+03 | 2.09E-12 | 9.45E+07 | 1.69E+07 | 9.85E+05 | 2.50E+03 | 1.00E+11 |
| | Median | 0.00E+00 | 1.75E+03 | 2.35E-12 | 3.41E+08 | 2.37E+07 | 9.88E+05 | 1.78E+05 | 5.31E+11 |
| | Worst | 0.00E+00 | 1.94E+03 | 2.39E+00 | 5.65E+08 | 3.43E+07 | 9.96E+05 | 3.06E+05 | 1.83E+12 |

|  | Mean | 0.00E+00 | 1.75E+03 | 6.02E-01 | 3.07E+08 | 2.41E+07 | 9.89E+05 | 1.46E+05 | 7.18E+11 |
|---|---|---|---|---|---|---|---|---|---|
|  | Stddev | 0.00E+00 | 1.06E+02 | 8.77E-01 | 1.47E+08 | 3.93E+06 | 2.67E+03 | 1.12E+05 | 5.42E+11 |
| **jDEsps** | Best | 1.95E-27 | 4.26E+01 | 4.97E-14 | 8.78E+08 | 2.05E+06 | 3.37E-01 | 6.98E+08 | 3.92E+12 |
|  | Median | 2.88E-23 | 9.15E+01 | 1.10E-13 | 2.56E+09 | 3.11E+06 | 7.07E-01 | 1.14E+09 | 4.76E+12 |
|  | Worst | 4.90E-22 | 7.65E+02 | 1.31E-12 | 2.72E+10 | 4.78E+06 | 2.24E+01 | 1.95E+09 | 2.27E+14 |
|  | Mean | 9.17E-23 | 1.69E+02 | 1.73E-13 | 4.41E+09 | 3.14E+06 | 4.55E+00 | 1.24E+09 | 1.76E+13 |
|  | Stddev | 1.37E-22 | 1.83E+02 | 2.55E-13 | 5.63E+09 | 6.38E+05 | 7.90E+00 | 4.06E+08 | 4.44E+13 |
| **MOS-CEC2013** | Best | 0.00E+00 | 7.06E+02 | 1.54E-12 | 4.77E+07 | 5.34E+06 | 6.16E+00 | 6.23E+01 | 5.49E+11 |
|  | Median | 0.00E+00 | 8.24E+02 | 1.69E-12 | 7.80E+07 | 6.95E+06 | 1.39E+05 | 1.10E+03 | 2.82E+12 |
|  | Worst | 0.00E+00 | 9.29E+02 | 1.95E-12 | 1.60E+08 | 9.40E+06 | 2.20E+05 | 5.30E+04 | 7.95E+12 |
|  | Mean | 0.00E+00 | 8.23E+02 | 1.69E-12 | 8.73E+07 | 6.89E+06 | 1.43E+05 | 4.65E+03 | 2.85E+12 |
|  | Stddev | 0.00E+00 | 4.69E+01 | 9.16E-14 | 3.11E+07 | 9.16E+05 | 6.86E+04 | 1.06E+04 | 1.44E+12 |
| **DECC-G** | Best | 8.47E-07 | 1.24E+03 | 1.05E-04 | 8.22E+10 | 5.73E+06 | 1.43E+05 | 4.55E+08 | 1.44E+15 |
|  | Median | 2.33E-06 | 1.31E+03 | 1.19E+00 | 2.08E+11 | 8.19E+06 | 1.77E+05 | 9.10E+08 | 6.10E+15 |
|  | Worst | 1.22E-05 | 1.38E+03 | 1.42E+00 | 4.12E+11 | 1.24E+07 | 2.28E+05 | 2.20E+09 | 1.63E+16 |
|  | Mean | 3.22E-06 | 1.31E+03 | 1.09E+00 | 2.16E+11 | 8.30E+06 | 1.74E+05 | 1.02E+09 | 6.94E+15 |
|  | Stddev | 2.83E-06 | 3.42E+01 | 3.54E-01 | 7.76E+10 | 1.60E+06 | 2.09E+04 | 4.89E+08 | 3.37E+15 |
| **CC-CMA-ES** | Best | 1.72E-09 | 1.15E+03 | 1.39E-13 | 9.86E+08 | 7.28E+14 | 1.18E+03 | 8.14E+04 | 6.83E+13 |
|  | Median | 5.56E-09 | 1.34E+03 | 1.49E-13 | 1.88E+09 | 7.28E+14 | 2.06E+05 | 1.36E+06 | 2.92E+14 |
|  | Worst | 7.52E-09 | 1.69E+03 | 1.63E-13 | 6.76E+09 | 7.28E+14 | 1.00E+06 | 1.04E+07 | 6.86E+14 |
|  | Mean | 5.36E-09 | 1.37E+03 | 1.50E-13 | 2.82E+09 | 7.28E+14 | 4.56E+05 | 2.26E+06 | 3.32E+14 |
|  | Stddev | 1.32E-09 | 1.38E+02 | 6.87E-15 | 1.84E+09 | 5.18E+06 | 4.56E+05 | 2.87E+06 | 1.74E+14 |

Table 4: Results of CEC 2010, CEC 2012, CEC 2013 and SOCO 2011 LSGO winners and proposed hybrid algorithms applied on CEC 2013 LSGO Benchmarks (F8, F9, F10, F11, F12, F13, F14)

| Algorithm | Metrics | F9 | F10 | F11 | F12 | F13 | F14 | F15 |
|---|---|---|---|---|---|---|---|---|
| **IMHS+MDE (Proposed)** | Best | 1.93E+08 | 91892953 | 4.8E+09 | 20302.52 | 3.45E+09 | 1.74E+10 | 13768197 |
|  | Median | 2.52E+08 | 92948392 | 1.72E+10 | 33536.33 | 4.27E+09 | 4.5E+10 | 36003393 |
|  | Worst | 3.04E+08 | 93626494 | 4.91E+10 | 83580.18 | 5.85E+09 | 6.02E+10 | 77715862 |
|  | Mean | 2.49E+08 | 92966830 | 2.11E+10 | 39065.84 | 4.39E+09 | 4.1E+10 | 37876036 |
|  | Stddev | 32949910 | 479080.6 | 1.26E+10 | 15600.81 | 5.78E+08 | 1.42E+10 | 17822924 |
| **M A-SW-Chains** | Best | 8.09E+07 | 8.63E+01 | 1.59E+08 | 1.08E+03 | 1.64E+07 | 1.11E+08 | 4.57E+06 |
|  | Median | 1.16E+08 | 3.32E+02 | 2.10E+08 | 1.24E+03 | 1.91E+07 | 1.47E+08 | 5.76E+06 |
|  | Worst | 1.54E+08 | 1.74E+05 | 2.42E+08 | 1.35E+03 | 2.55E+07 | 1.70E+08 | 1.16E+07 |
|  | Mean | 1.14E+08 | 3.66E+04 | 2.10E+08 | 1.23E+03 | 1.98E+07 | 1.45E+08 | 5.90E+06 |
|  | Stddev | 2.05E+07 | 6.17E+04 | 2.43E+07 | 8.32E+01 | 2.30E+06 | 1.69E+07 | 1.36E+06 |
| **2S-Ensemble** | Best | 1.05E+08 | 9.22E+06 | 8.26E+07 | 1.07E+03 | 6.51E+07 | 3.61E+07 | 1.87E+06 |
|  | Median | 1.34E+08 | 1.44E+07 | 2.45E+08 | 1.66E+03 | 2.14E+08 | 5.68E+07 | 2.37E+06 |
|  | Worst | 1.55E+08 | 2.20E+07 | 3.35E+08 | 2.02E+03 | 7.31E+08 | 9.73E+07 | 2.61E+06 |
|  | Mean | 1.31E+08 | 1.43E+07 | 2.38E+08 | 1.70E+03 | 2.80E+08 | 6.09E+07 | 2.33E+06 |
|  | Stddev | 1.51E+07 | 2.87E+06 | 6.45E+07 | 2.20E+02 | 1.75E+08 | 1.75E+07 | 1.74E+05 |
| **MOS-SOCO2011** | Best | 5.02E+08 | 1.80E+07 | 3.83E+08 | 1.91E-02 | 2.97E+08 | 1.29E+08 | 3.40E+07 |
|  | Median | 9.16E+08 | 8.78E+07 | 1.35E+09 | 7.01E+01 | 8.37E+08 | 2.30E+09 | 1.21E+08 |
|  | Worst | 1.09E+09 | 8.97E+07 | 5.74E+11 | 2.97E+02 | 3.65E+09 | 6.55E+10 | 2.66E+08 |
|  | Mean | 8.97E+08 | 6.65E+07 | 4.01E+10 | 8.63E+01 | 1.13E+09 | 6.89E+09 | 1.31E+08 |
|  | Stddev | 1.39E+08 | 2.91E+07 | 1.23E+11 | 7.74E+01 | 7.74E+08 | 1.41E+10 | 6.02E+07 |
| **jDElscop** | Best | 2.02E+08 | 5.61E+02 | 1.81E+10 | 9.70E+02 | 2.17E+10 | 2.53E+11 | 2.58E+07 |
|  | Median | 3.67E+08 | 1.01E+03 | 1.22E+11 | 9.74E+02 | 3.48E+10 | 4.89E+11 | 3.43E+07 |
|  | Worst | 4.46E+08 | 1.30E+03 | 2.15E+11 | 1.03E+03 | 4.86E+10 | 6.58E+11 | 7.28E+07 |
|  | Mean | 3.66E+08 | 1.06E+03 | 1.15E+11 | 9.91E+02 | 3.49E+10 | 4.59E+11 | 3.98E+07 |
|  | Stddev | 5.57E+07 | 1.72E+02 | 5.86E+10 | 2.70E+01 | 5.58E+09 | 1.16E+11 | 1.50E+07 |
| **GaDE** | Best | 4.84E+08 | 1.52E+07 | 1.32E+11 | 1.03E+03 | 2.12E+10 | 3.13E+11 | 7.76E+07 |
|  | Median | 6.78E+08 | 2.29E+07 | 2.41E+11 | 1.11E+03 | 2.87E+10 | 4.19E+11 | 1.09E+08 |
|  | Worst | 7.74E+08 | 3.58E+07 | 3.51E+11 | 1.33E+03 | 3.54E+10 | 6.57E+11 | 1.33E+08 |
|  | Mean | 6.62E+08 | 2.35E+07 | 2.40E+11 | 1.12E+03 | 2.88E+10 | 4.21E+11 | 1.07E+08 |
|  | Stddev | 7.44E+07 | 4.59E+06 | 5.96E+10 | 6.90E+01 | 3.41E+09 | 7.13E+10 | 1.43E+07 |
|  | Best | 1.00E+09 | 8.81E+07 | 1.85E+07 | 1.50E-02 | 4.70E+06 | 7.96E+06 | 4.89E+06 |
|  | Median | 1.65E+09 | 9.01E+07 | 2.53E+07 | 1.68E+02 | 5.90E+06 | 4.80E+07 | 7.25E+06 |
| **MOS-CEC2012** | Worst | 2.33E+09 | 9.07E+07 | 3.89E+07 | 8.19E+02 | 6.84E+06 | 6.48E+07 | 1.04E+07 |
| | Mean | 1.65E+09 | 9.00E+07 | 2.71E+07 | 2.21E+02 | 5.84E+06 | 3.78E+07 | 7.63E+06 |
| | Stddev | 3.47E+08 | 5.04E+05 | 5.19E+06 | 2.29E+02 | 6.31E+05 | 2.01E+07 | 1.53E+06 |
| **jDEsps** | Best | 2.21E+08 | 8.71E+02 | 1.03E+10 | 3.34E+02 | 2.29E+10 | 7.54E+10 | 2.94E+07 |
| | Median | 2.74E+08 | 9.21E+02 | 2.74E+10 | 7.89E+02 | 3.12E+10 | 1.80E+11 | 4.59E+07 |
| | Worst | 3.21E+08 | 1.91E+03 | 5.20E+10 | 1.71E+03 | 3.82E+10 | 4.0BE+11 | 7.49E+07 |
| | Mean | 2.71E+08 | 1.05E+03 | 2.90E+10 | 8.51E+02 | 3.07E+10 | 1.93E+11 | 4.58E+07 |
| | Stddev | 2.94E+07 | 2.43E+02 | 1.13E+10 | 3.64E+02 | 4.16E+09 | 8.24E+10 | 1.12E+07 |
| **MOS-CEC2013** | Best | 2.33E+08 | 4.44E+02 | 8.89E+06 | 4.40E-03 | 3.08E+05 | 7.83E+06 | 1.42E+06 |
| | Median | 4.18E+08 | 1.17E+06 | ==1.71E+07== | ==1.56E+01== | ==1.02E+06== | ==1.28E+07== | ==1.68E+06== |
| | Worst | 5.01E+08 | 1.23E+06 | 2.85E+07 | 7.48E+02 | 2.22E+06 | 2.09E+07 | 2.03E+06 |
| | Mean | 3.99E+08 | 9.38E+05 | 1.73E+07 | 8.13E+01 | 1.00E+06 | 1.24E+07 | 1.71E+06 |
| | Stddev | 6.26E+07 | 4.79E+05 | 5.04E+06 | 1.57E+02 | 5.53E+05 | 2.86E+06 | 1.44E+05 |
| **DECC-G** | Best | 3.95E+08 | 7.87E+06 | 7.96E+09 | 3.59E+03 | 5.15E+09 | 6.77E+10 | 9.54E+06 |
| | Median | 5.32E+08 | 2.15E+07 | 1.25E+11 | 4.20E+03 | 8.66E+09 | 1.23E+11 | 1.19E+07 |
| | Worst | 7.82E+08 | 5.19E+07 | 2.32E+11 | 7.24E+03 | 1.75E+10 | 2.32E+11 | 1.42E+07 |
| | Mean | 5.47E+08 | 2.43E+07 | 1.21E+11 | 4.53E+03 | 9.40E+09 | 1.36E+11 | 1.17E+07 |
| | Stddev | 8.94E+07 | 9.75E+06 | 6.03E+10 | 9.48E+02 | 3.15E+09 | 4.79E+10 | 1.18E+06 |
| **CC-CMA-ES** | Best | 1.76E+08 | 1.32E+05 | 4.17E+07 | 9.78E+02 | 1.79E+07 | 2.27E+07 | 2.31E+07 |
| | Median | 3.57E+08 | 7.48E+05 | 9.79E+07 | 1.03E+03 | 4.12E+08 | 3.66E+07 | 3.05E+07 |
| | Worst | 6.77E+08 | 9.03E+07 | 5.04E+08 | 2.23E+03 | 1.90E+10 | 5.78E+09 | 4.29E+07 |
| | Mean | 3.82E+08 | 4.51E+06 | 1.24E+08 | 1.33E+03 | 1.80E+09 | 3.58E+08 | 3.13E+07 |
| | Stddev | 1.48E+08 | 1.79E+07 | 9.88E+07 | 4.78E+02 | 4.02E+09 | 1.15E+09 | 5.30E+06 |

Statistical analysis is performed to test the effectiveness of the proposed algorithm. Table 5 presents how many times numerical rank 1 was obtained by all the algorithms (nBest) over 15 functions. In Table 5, the nBest value for the proposed algorithm is one and for the MOS-CEC2013 is eight, which is the highest.

In order to confirm whether the superior performance of algorithms is due to the strength of the algorithms or merely due to chance of happening. Therefore, a statistical ranking of all the algorithms is performed using two multiple comparison statistical tests viz., Friedman's test and Quade's test. Friedman's test ranked the proposed algorithm at an an absolute rank of 7.87.. The lowest absolute rank for Friedman's test is 3.23 for MOS-CEC2013. Now to consider variability in ranking, Quade's test was performed. Quade's test gave a absolute rank of 6.65 for the proposed algorithm and a lowest absolute rank of 2.65. There is no absolute rank of 1 in both tests which shows that no algorithm is a clear winner. Quade's test also suggests the same as Friedman's test from p-values that there is a difference among the algorithms, which can be seen in the rankings of the tests presented in Table 5. Therefore, from the above two statistical tests we conclude that our proposed algorithms is in absolute rank of 7.87 and 6.65 which is close to 8 or 7 out of 11, which is creditworthy performance.

Table 5: Number of best values and ranking based on statistical tests

| Algorithm | nBest | Friedman's Test (Ranking) | Quade's Test (Ranking) |
|---|---|---|---|
| IMHS+MDE | 1 | 7.87 | 6.65 |
| M A-SW-Chains | 2 | 4.47 | 3.98 |
| 2S-Ensemble | 1 | 4.9333 | 4.9625 |
| MOS-SOCO2011 | 2 | 6.7667 | 7.7125 |
| jDElscop | 2 | 6.0667 | 6.9375 |
| GaDE | 2 | 7.7 | 8.5125 |
| MOS-CEC2012 | 2 | 5.5 | 4.5625 |
| jDEsps | 0 | 5.4667 | 5.9417 |
| MOS-CEC2013 | 8 | 3.2333 | 2.6542 |
| DECC-G | 0 | 8.4 | 8.5583 |
| CC-CMA-ES | 0 | 5.6 | 5.5333 |
| Statistic Value | | **33.50** | **4.71** |
| p-value | | **0.0001946** | **1.275e-05** |

Apart from statistical ranking, a score-based ranking of algorithms was also performed. As proposed in CEC 2015 LSGO, the ranking was performed based on a median value of fitness value over 25 runs for each benchmark function for all algorithms. The median rankings are presented in Tables 6-7.

Table 6: Ranking based on Medians (As performed in CEC 2015 LSGO) for benchmarks F1 to F10

| Algorithm | F1 | R1 | F2 | R2 | F3 | R3 | F4 | R4 | F5 | R5 | F6 | R6 | F7 | R7 | F8 | R8 |
|---|---|---|---|---|---|---|---|---|---|---|---|---|---|---|---|---|
| IMHS+MDE | 4595495 | 11 | 1564.223359 | 10 | 20.68593502 | 11 | 9746262377 | 7 | 932950.4438 | 1 | 1059663.104 | 11 | 1.42E+08 | 7 | 1.53161E+14 | 6 |
| M A-SW-Chains | 6.15E-13 | 8 | 1.13E+03 | 7 | 6.79E-13 | 7 | 2.70E+09 | 5 | 1.98E+06 | 3 | 6.24E+02 | 3 | 3.99E+06 | 5 | 4.65E+13 | 4 |
| 2S-Ensemble | 0.00E+00 | 1 | 9.95E-01 | 2 | 3.98E-13 | 6 | 1.13E+10 | 9 | 1.28E+06 | 2 | 2.00E+05 | 6 | 1.54E+06 | 4 | 3.94E+14 | 9 |
| MOS-SOCO2011 | 0.00E+00 | 1 | 1.96E+01 | 4 | 0.00E+00 | 1 | 1.20E+10 | 10 | 1.09E+07 | 9 | 9.85E+05 | 8 | 1.23E+07 | 6 | 1.13E+15 | 10 |
| jDElscop | 0.00E+00 | 1 | 9.95E-01 | 2 | 2.77E-13 | 5 | 7.18E+09 | 6 | 4.83E+06 | 5 | 1.99E-01 | 1 | 3.17E+09 | 11 | 1.05E+14 | 5 |
| GaDE | 0.00E+00 | 1 | 0.00E+00 | 1 | 2.56E-13 | 4 | 1.01E+10 | 8 | 1.02E+07 | 8 | 1.03E+06 | 10 | 2.68E+09 | 10 | 1.57E+14 | 7 |
| MOS-CEC2012 | 0.00E+00 | 1 | 1.75E+03 | 11 | 2.35E-12 | 9 | 3.41E+08 | 2 | 2.37E+07 | 10 | 9.88E+05 | 9 | 1.78E+05 | 2 | 5.31E+11 | 1 |
| jDEsps | 2.88E-23 | 7 | 9.15E+01 | 5 | 1.10E-13 | 2 | 2.56E+09 | 4 | 3.11E+06 | 4 | 7.07E-01 | 2 | 1.14E+09 | 9 | 4.76E+12 | 3 |
| MOS-CEC2013 | 0.00E+00 | 1 | 8.24E+02 | 6 | 1.69E-12 | 8 | 7.80E+07 | 1 | 6.95E+06 | 6 | 1.39E+05 | 4 | 1.10E+03 | 1 | 2.82E+12 | 2 |
| DECC-G | 2.33E-06 | 10 | 1.31E+03 | 8 | 1.19E+00 | 10 | 2.08E+11 | 11 | 8.19E+06 | 7 | 1.77E+05 | 5 | 9.10E+08 | 8 | 6.10E+15 | 11 |
| CC-CMA-ES | 5.56E-09 | 9 | 1.34E+03 | 9 | 1.49E-13 | 3 | 1.88E+09 | 3 | 7.28E+14 | 11 | 2.06E+05 | 7 | 1.36E+06 | 3 | 2.92E+14 | 8 |

Table 7: Ranking based on Medians (As performed in CEC 2015 LSGO) for benchmarks F11 to F20

| Algorithm | F9 | R9 | F10 | R10 | F11 | R11 | F12 | R12 | F13 | R13 | F14 | R14 | F15 | R15 |
|---|---|---|---|---|---|---|---|---|---|---|---|---|---|---|
| IMHS+MDE | 2.52E+08 | 3 | 92948392 | 11 | 1.72E+10 | 7 | 33536.33 | 11 | 4.27E+09 | 7 | 4.5E+10 | 7 | 36003393 | 8 |
| M A-SW-Chains | 1.16E+08 | 1 | 3.32E+02 | 1 | 2.10E+08 | 4 | 1.24E+03 | 8 | 1.91E+07 | 3 | 1.47E+08 | 5 | 5.76E+06 | 3 |
| 2S-Ensemble | 1.34E+08 | 2 | 1.44E+07 | 6 | 2.45E+08 | 5 | 1.66E+03 | 9 | 2.14E+08 | 4 | 5.68E+07 | 4 | 2.37E+06 | 2 |
| MOS-SOCO2011 | 9.16E+08 | 10 | 8.78E+07 | 9 | 1.35E+09 | 6 | 7.01E+01 | 2 | 8.37E+08 | 6 | 2.30E+09 | 6 | 1.21E+08 | 11 |
| jDElscop | 3.67E+08 | 6 | 1.01E+03 | 3 | 1.22E+11 | 9 | 9.74E+02 | 5 | 3.48E+10 | 11 | 4.89E+11 | 11 | 3.43E+07 | 7 |
| GaDE | 6.78E+08 | 9 | 2.29E+07 | 8 | 2.41E+11 | 11 | 1.11E+03 | 7 | 2.87E+10 | 9 | 4.19E+11 | 10 | 1.09E+08 | 10 |
| MOS-CEC2012 | 1.65E+09 | 11 | 9.01E+07 | 10 | 2.53E+07 | 2 | 1.68E+02 | 3 | 5.90E+06 | 2 | 4.80E+07 | 3 | 7.25E+06 | 4 |
| jDEsps | 2.74E+08 | 4 | 9.21E+02 | 2 | 2.74E+10 | 8 | 7.89E+02 | 4 | 3.12E+10 | 10 | 1.80E+11 | 9 | 4.59E+07 | 9 |
| MOS-CEC2013 | 4.18E+08 | 7 | 1.17E+06 | 5 | 1.71E+07 | 1 | 1.56E+01 | 1 | 1.02E+06 | 1 | 1.28E+07 | 1 | 1.68E+06 | 1 |
| DECC-G | 5.32E+08 | 8 | 2.15E+07 | 7 | 1.25E+11 | 10 | 4.20E+03 | 10 | 8.66E+09 | 8 | 1.23E+11 | 8 | 1.19E+07 | 5 |
| CC-CMA-ES | 3.57E+08 | 5 | 7.48E+05 | 4 | 9.79E+07 | 3 | 1.03E+03 | 6 | 4.12E+08 | 5 | 3.66E+07 | 2 | 3.05E+07 | 6 |

Formula one (F1) based scoring (see Table 8) was performed after the algorithms were ranked according to median. Any function that is ranked above ten is given a zero score. A summation, (all the scores of the benchmark functions for a particular algorithm) was performed, and later ranking was done based on these scores. The highest F1 score has the lowest rank (1), and the lowest F1 score has the last rank.

Table 8: F1 racing points table for score calculation

| Formula One (F1) Points Table ||
|---|---|
| 1st place | 25 points |
| 2nd place | 18 points |
| 3rd place | 15 points |
| 4th place | 12 points |
| 5th place | 10 points |
| 6th place | 8 points |
| 7th place | 6 points |
| 8th place | 4 points |
| 9th place | 2 points |
| 10th place | 1 point |
| 11th place onwards | No points |

To better understand in which functionality grouping (separable, partially separable or non-separable) the proposed algorithm is performing well, we have also calculated the F1 score for each algorithm based on functionality grouping. The F1 scores have been presented in Table 9. The F1 score for functionality grouping suggests that the proposed algorithm is performing moderately well for partially separable functions, for overlapping functions, and for non-separable functions, not performing well for separable functions.

In this score-based ranking, MOS-CEC2013 algorithm, which is the winner of all the competitions, is higher in the ranking than the proposed algorithm. Our proposed hybrid is better than MOS-SOCO2011 (Rank 1 of SOCO 2011 Competition), GaDE (Rank 3 of SOCO 2011 Competition), and DECC-G (Rank 2 of CEC 2013 LSGO competition).

Table 9: F1 Score based on Median Ranking (As given in CEC 2015 LSGO)

| Functions | Separable Functions | | Partially Separable Functions and Overlapping Functions | | | | | | Non-separable Functions | |
|---|---|---|---|---|---|---|---|---|---|---|
| | (F1, F2, F3) | | (F4, F5, F6, F7) | | (F8, F9, F10, F11) | | (F12, F13, F14) | | (F15) | |
| Algorithm | S1 | R1 | S2 | R2 | S3 | R3 | S4 | R4 | S5 | R5 |
| IMHS+MDE | 1 | 11 | 37 | 7 | 29 | 8 | 12 | 8 | 4 | 8 |
| M A-SW-Chains | 16 | 9 | 50 | 2 | 74 | 1 | 29 | 5 | 15 | 3 |
| 2S-Ensemble | 51 | 4 | 40 | 5 | 38 | 5 | 26 | 6 | 18 | 2 |
| MOS-SOCO2011 | 62 | 1 | 15 | 10 | 12 | 9 | 34 | 4 | 0 | 11 |
| jDElscop | 53 | 3 | 43 | 4 | 35 | 7 | 10 | 9 | 6 | 7 |
| GaDE | 62 | 1 | 10 | 11 | 12 | 9 | 9 | 10 | 1 | 10 |
| MOS-CEC2012 | 27 | 7 | 39 | 6 | 44 | 4 | 48 | 2 | 12 | 4 |
| jDEsps | 34 | 6 | 44 | 3 | 49 | 3 | 15 | 7 | 3 | 9 |
| MOS-CEC2013 | 37 | 5 | 70 | 1 | 59 | 2 | 75 | 1 | 25 | 1 |
| DECC-G | 6 | 10 | 20 | 9 | 11 | 11 | 9 | 10 | 10 | 5 |
| CC-CMA-ES | 19 | 8 | 36 | 8 | 37 | 6 | 36 | 3 | 8 | 6 |

## 8. Conclusions and Future Directions

We conclude that we proposed a novel hybrid meta-heuristic and tested its effectiveness on CEC 2013 LSGO benchmarks. Statistical tests such as Quade's and Friedman's tests were conducted on the compared algorithms. Apart from these statistical tests, F1 based scoring was also performed on the compared algorithms. The experimental results consistently show that our proposed hybrid meta-heuristic performs statistically on par with some algorithms in a few problems, while it turned out to be the best in a couple of problems.